\documentclass[runningheads]{llncs}

\usepackage[mobile]{eccv}

\usepackage{eccvabbrv}

\usepackage{graphicx}
\usepackage{booktabs}
\usepackage{multirow}
\usepackage[table]{xcolor}
\usepackage{wrapfig}

\usepackage[accsupp]{axessibility}  

\usepackage{hyperref}

\usepackage{orcidlink}

\begin{document}

\newcommand{\method}{WorldWander\xspace}

\title{\method: Bridging Egocentric and Exocentric Worlds in Video Generation}
\titlerunning{Bridging Egocentric and Exocentric Worlds in Video Generation}

\author{
Quanjian Song\inst{2}$^{*}$ \and
Yiren Song\inst{1}$^{*}$ \and
Kelly Peng\inst{2} \and
Yuan Gao\inst{2} \and
Mike Zheng Shou\inst{1}$^{\dagger}$
}

\authorrunning{Q. Song et al.}

\institute{
\mbox{$^{1}$Show Lab, National University of Singapore, Singapore} \quad
\mbox{$^{2}$First Intelligence}
\\
\url{https://github.com/showlab/WorldWander}
}

\maketitle

\begingroup
\renewcommand{\thefootnote}{}
\footnotetext{
\textsuperscript{$*$}Equal contribution.
\quad
\textsuperscript{$\dagger$}Corresponding author.
}
\endgroup

\begin{center}
  \includegraphics[width=\textwidth]{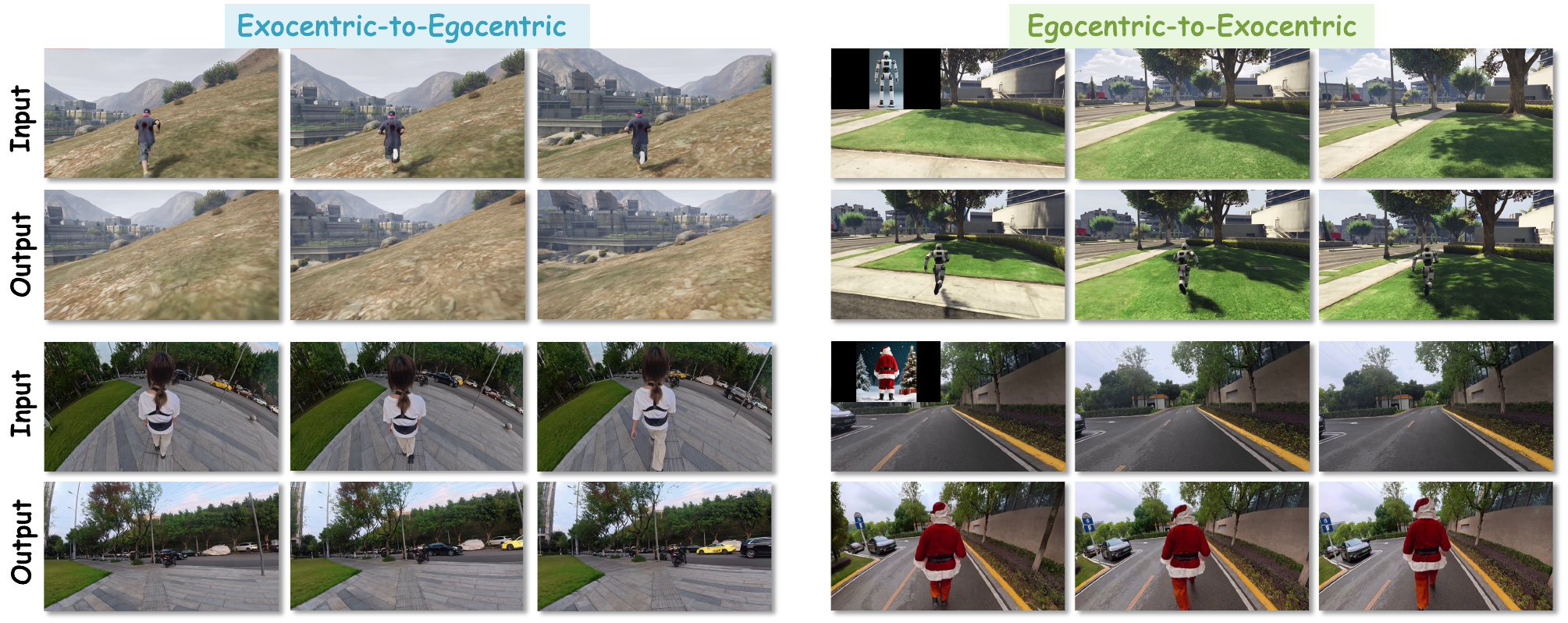}
  \captionof{figure}{
  Gallery of our \method. It bridges the egocentric and exocentric worlds in video generation, enabling character-centric and immersive navigation.
}
  \label{fig:teaser}
\end{center}

\begin{abstract}
Recent advances in video world models enable interactive environments with free navigation, making translation between first-person (egocentric) and third-person (exocentric) perspectives increasingly important.
However, existing studies focus on unidirectional exocentric-to-egocentric translation, overlooking reference-guided exocentric perspective synthesis. This capability is crucial for gaming and embodied AI applications.
Motivated by this, we present WorldWander, an in-context learning framework tailored for translating between egocentric and exocentric worlds in video generation.
Building upon advanced video diffusion transformers, WorldWander integrates (i) In-Context Perspective Alignment and (ii) Collaborative Position Encoding to model cross-view synchronization and character consistency.
To support our task, we curate EgoExo-8K, a dynamic and scene-rich dataset containing synchronized egocentric–exocentric triplets from both synthetic and real-world scenarios.
Experiments demonstrate that WorldWander achieves superior perspective synchronization, character consistency, and generalization, setting a new benchmark for egocentric-exocentric video translation.
\end{abstract}

\section{Introduction}
\label{sec:intro}
Driven by the advancements in diffusion models~\cite{song2020denoising,song2020score,lipman2022flow}, video generation has attracted increasing attention.
Early research primarily focus on producing temporally coherent videos with controllable attributes such as camera motion~\cite{CameraCtrl,MotionCtrl}, scene~\cite{xie2025geometry,song2026scenedecorator}, and style~\cite{song2025univst, song2024processpainter,song2026makeanything}.
Building upon these advances, subsequent works~\cite{Recapture,TrajectoryCrafter,Recammaster,Reangle-a-video,chen2026transanimate} explore perspective re-orientation, aiming to generate videos from source perspectives to target perspectives.
  More recently, emerging world models~\cite{bar2025navigation,he2025matrix,li2025hunyuan} further extend this paradigm by constructing interactive virtual environments that enable free navigation.
As users naturally navigate these environments, the ability to seamlessly switch between first-person (egocentric) and third-person (exocentric) perspectives becomes increasingly important.
However, existing approaches~\cite{4Diff,EgoExo4D,Egoexo-gen} are largely limited to one-directional exocentric-to-egocentric translation and overlook the inverse problem of reference-guided exocentric perspective synthesis. Such capability is essential for applications including gaming production, embodied AI, and VR, where flexible perspective switching enables immersive and character-centric exploration in the worlds.

In this work, we first formulate the task of \textbf{Egocentric–Exocentric Perspective Translation}, which aims to seamlessly translate video across egocentric and exocentric perspectives during dynamic scene navigation.
Accordingly, we identify three primary challenges inherent to this setting:
(i) \textit{Perspective Synchronization.} Existing unidirectional egocentric-to-exocentric methods~\cite{4Diff,EgoExo4D,Egoexo-gen} are largely confined to static cameras. In our dynamic scene navigation, they struggle to maintain synchronization of key elements (\emph{e.g.}, appearance, motion rhythm, and environmental dynamics) between these two perspectives.
(ii) \textit{Character Consistency.} The novel exo-to-ego translation task requires generating customized exocentric views conditioned on an egocentric input and a reference image. Ensuring character consistency is non-trivial, as the model must associate the first-person observations with the corresponding third-person identity.
(iii) \textit{Data Scarcity.} While prior works~\cite{{EgoExo4D,H2O,TACO}} provide paired egocentric–exocentric datasets, they are heavily restricted to static cameras. Such static setups fail to support our ``WorldWander'' vision, where a personalized character can freely navigate via dynamic environments with seamless video perspective translation.

To address these issues above, we propose \textbf{\method}, an in-context learning framework tailored for translating between egocentric and exocentric worlds in video generation (see \cref{fig:teaser}).
Unlike traditional novel-view synthesis methods relying on 3D geometry or explicit camera pose, \method performs in-context learning directly from triplet datasets.
Specifically, it is built upon the state-of-the-art video diffusion transformer Wan2.2~\cite{wan2025wan} and integrates two core components for efficient fine-tuning:
(i) \textbf{In-Context Perspective Alignment}, which transforms the conditional and target videos into a shared latent space, followed by collaborative attention to model cross-view alignment and character consistency; and
(ii) \textbf{Collaborative Position Encoding}, which enables the conditional and target latents to share the unified positional embedding along the temporal dimension, thereby further enhancing cross-view consistency.

To tackle data scarcity, we carefully design a two-player protocol for recording synthetic scenes in GTA-5, alongside a dual-camera setup for capturing real-world scenarios.
Ultimately, we curate \textbf{EgoExo-8K}, a dynamic and scene-rich dataset featuring synchronized egocentric and exocentric perspectives. It comprises approximately 4,000 synthetic and 4,000 real-world triplets, each containing an egocentric video, its exocentric counterpart, and a reference image. 
Notably, unlike previous datasets~\cite{EgoExo4D,H2O,TACO} confined to static camera viewpoints, EgoExo-8K features dynamic scene navigation in open-world environments.
Comprehensive experiments demonstrate that WorldWander achieves high cross-view consistency and robust generalization in both transformations (ego$\rightarrow$exo and exo$\rightarrow$ego), significantly outperforming all existing baselines.

In summary, the main contributions of this paper are threefold:
\begin{itemize}
    \item We formulate the \textit{Egocentric–Exocentric Video Translation} tasks, where the novel exo-to-ego translation enables customized egocentric synthesis conditioned on a reference image, which is unexplored in prior works.
    \item We propose an in-context learning framework that is independent of camera pose and depth map, featuring \textit{In-Context Perspective Alignment} and \textit{Collaborative Position Encoding} for geometry-free cross-view translation.
    \item We carefully curate \textit{EgoExo-8K}, the first dynamic and open-world scene wandering benchmark for the egocentric–exocentric video translation, on which our \method achieves state-of-the-art performance across baselines.
\end{itemize}

\section{Related Work}
\label{sec:relatedwork}

\subsection{Exo-to-Ego Perspective Translation}
Paired exocentric-egocentric data like EgoExo4D~\cite{EgoExo4D}, H2O~\cite{H2O}, and TACO~\cite{TACO} have driven recent progress in exocentric-to-egocentric view translation. Essentially, it involves synthesizing first-person (egocentric) perspective from third-person (exocentric) perspective while maintaining spatio-temporal consistency.
Early works~\cite{luo2024intention,park2025egoworld,luo2024put} fuse exocentric and egocentric latents via channel-wise concatenation.
Given that this approach suffers from information loss and sluggish convergence, later methods~\cite{liu2024exocentric,4Diff} shift towards cross-attention mechanisms to inject exocentric features.
Subsequently, several works propose task variants that rely on auxiliary inputs for more accurate generation. For instance, Exo2Ego-V~\cite{ExoEgoV} leverages four exocentric views, while EgoExo-Gen~\cite{Egoexo-gen} requires the first egocentric frame as an additional condition.
However, these works typically focus on unidirectional ego-to-exo translation restricted to static viewpoints.
In contrast, our work tackle bidirectional ego-exo translation in dynamic wandering scenarios with freely moving cameras. 
Furthermore, for the novel exo-to-ego translation, our method enables customized egocentric perspective synthesis using a reference character, which remains unexplored in prior works.

\subsection{Viewpoint-Control in Video Generation}
Driven by industrial demand, research on controllability in diffusion models has grown rapidly \cite{song2025layertracer,zhang2025easycontrol,song2026omniconsistency,gong2026relationadapter, jiang2025personalized, wang2026diffdecompose,lu2026easytext, shi2025fonts,qu2025drag,song2026fashionchameleon}, providing a solid foundation for viewpoint-controllable video generation.
Early on, some methods~\cite{ma2024trailblazer,MotionCtrl,CameraCtrl,zhang2025tora} typically train an additional encoder to accept camera parameters and produce videos with camera movements; while other approaches~\cite{CamTrol,song2025lightmotion,MotionBooth} explore training-free techniques to simulate camera motion.
Over time, some studies such as TrajectoryCrafter~\cite{TrajectoryCrafter}, Recammaster~\cite{Recammaster}, and Reangle-a-video~\cite{Reangle-a-video} have explored the task of perspective re-orientation. Given a reference video, these methods aim to transform it to the target perspective while maintaining spatial-temporal consistency.
The emergence of world models such as Genie3, Matrix-Game 2.0~\cite{he2025matrix}, and Hunyuan-GameCraft~\cite{li2025hunyuan} has enabled video diffusion models to construct virtual worlds that users can navigate through camera movements.
Despite these advances in video generation, existing approaches still struggle to model egocentric–exocentric transformations, which are the focus of this work.

\subsection{Video-to-Video Translation}
As a sub-task of controllable video generation, video-to-video translation aims to take an additional video as the condition and transform it into another video.
On one hand, different studies explore this task from various dimensions: UniVST~\cite{song2025univst}, StyleCrafter~\cite{liu2024stylecrafter}, and StyleMaster~\cite{ye2025stylemaster} focus on stylizing conditional videos into specific artistic forms; FlowV2V~\cite{wang2026consistent} and Rerender-A-Video~\cite{yang2023rerender} explore video editing for controllable outcomes.
On the other hand, several works such as AnyV2V~\cite{ku2024anyv2v}, I2VEdit~\cite{ouyang2024i2vedit}, and VACE~\cite{jiang2025vace} have sought to unify different forms of Video-to-Video Generation within a single framework. These frameworks are capable of performing diverse tasks, including style transfer \cite{huang2025photodoodle,song2026vista}, inpainting \cite{gao2026pai}, outpainting, and others \cite{hu2024animate,zhang2024ssr,song2026streamingeffect}.
However, none of these approaches explores the transformations between egocentric and exocentric perspectives. Our work aims to bridge this gap in the field.

\section{Preliminary}

\noindent
\textbf{Video Diffusion Transformer.}
The state-of-the-art video diffusion transformer~\cite{peebles2023scalable} is typically composed of a pair of variational encoder $\mathcal{E}$ and decoder $\mathcal{D}$, as well as a transformer-based prediction network $v_{\theta}$.
During training, the encoder $\mathcal{E}$ will transform $F$ frames of the target video into $f$ frames of the latent representation $z_{0}^{1:f}$, where $f = \frac{F - 1}{4} + 1$.
According to the rectified-flow~\cite{lipman2022flow}, the forward process is defined as a straight path from the data distribution to a standard normal distribution. This process is formulated below:
\begin{equation}
    z_t^{1:f} = (1 - t) \cdot z_0^{1:f} + t \cdot \epsilon^{1:f},
\label{eq:1}
\end{equation}
where $\epsilon^{1:f} \in \mathcal{N}(0, I)$ and $t$ is a random timestep.
Given the noisy latent $z_t^{1:f}$, we utilize the network $v_{\theta}$ to regress the vector field via flow matching~\cite{lipman2022flow} loss:
\begin{equation}
    \min_{\theta} \mathbb{E}_{t \sim \mathcal{U}(0,1)} \| v_{\theta}(z_{t}^{1:f}, t, c) - v \|_2^2,
\label{eq:2}
\end{equation}
where $v = \epsilon^{1:f} - z_0^{1:f}$ denotes the target vector field, and $c$ represents the additional conditioning signal (\emph{e.g.}, text, image, video).

During inference, the trained $v_{\theta}$ will iteratively map samples from Gaussian distribution back to specific data distribution, by solving the ODE below:
\begin{equation}
    dz_t^{1:f} = v_{\theta}(z_{t}^{1:f}, t, c) dt.
\end{equation}
Finally, the decoder $\mathcal{D}$ will reconstruct final video from the denoised latent.

\noindent
\textbf{Rotary Position Encoding.}
In advanced video generation models~\cite{li2025hunyuan,wan2025wan,yang2024cogvideox}, rotary position embeddings are typically applied within the attention process, thereby distinguishing tokens across spatial and temporal dimensions.

In detail, customized positional embeddings $\mathcal{P}$ will be injected into the query and key representations ($Q$ and $K$) before the attention computation:
\begin{equation}
    \begin{aligned}
    Q[i,j,k] &\gets Q[i,j,k] \cdot \mathcal{P}[i,j,k], \\
    K[i,j,k] &\gets K[i,j,k] \cdot \mathcal{P}[i,j,k],
    \end{aligned}
\label{eq:rope}
\end{equation}
where $i$, $j$, and $k$ indicate the indices along the height, width, and frame.
We elaborate in the methodology section on how the positional embeddings $\mathcal{P}$ are constructed to adapt for our egocentric–exocentric video translation.

\begin{figure*}[t]
    \centering
    \includegraphics[width=0.98\linewidth]{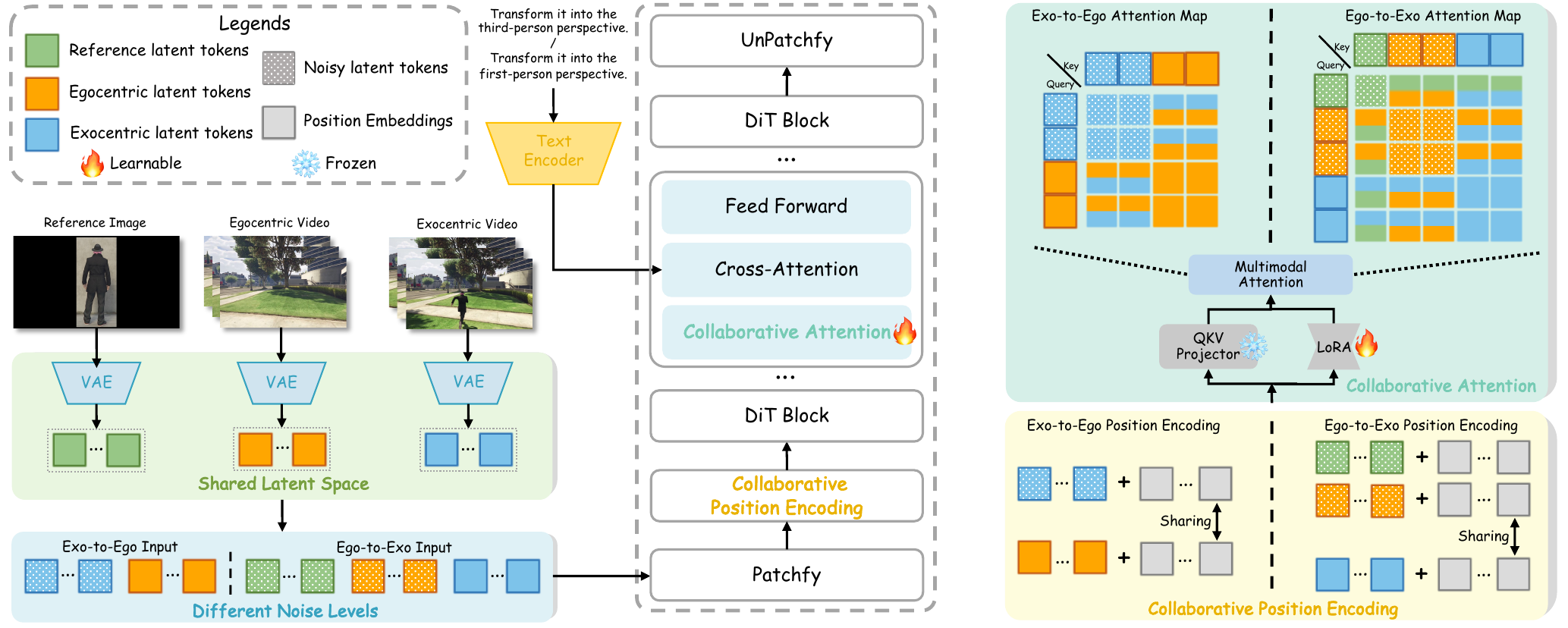}
    \caption{
    Overall pipeline of \method. The backbone Wan2.2-5B~\cite{wan2025wan} is fine-tuned with proposed \textit{In-Context Perspective Alignment} (Shared Latent Space, Different Noise Levels, and Collaborative Attention) as well as \textit{Collaborative Position Encoding}.
    }
    \label{fig:overall_pipeline}
\end{figure*}

\section{Methodology}
In this work, we propose \textbf{\method}, an in-context learning framework that bridges first-person (egocentric) and third-person (exocentric) perspectives in video generation, enabling immersive and character-centric world exploration. The overall pipeline is shown in \cref{fig:overall_pipeline}.
First, we describe our \textit{Task Formulation} in \cref{sec:4.1}.
Then, we present the \textit{In-Context Perspective Alignment} paradigm in \cref{sec:4.2} and the \textit{Collaborative Position Encoding} strategy in \cref{sec:4.3}, which model the perspective correspondence and character consistency without auxiliary networks.
Subsequently, we provide details of \textit{Efficient Fine-Tuning via LoRA} in \cref{sec:4.4}.
Finally, we present curated \textit{EgoExo-8K} dataset in \cref{sec:4.5}.

\subsection{Task Formulation}
\label{sec:4.1}
This paper tackles a challenging yet applicable task of egocentric-exocentric video translation.
In detail, for the triplet sample $\langle \mathcal{V}^{\text{ego}}, \mathcal{V}^{\text{exo}}, \mathcal{I}^{\text{ref}} \rangle$, egocentric-exocentric translation consists of two sub-tasks:
(i) \textit{Exocentric-to-egocentric translation}, where the exocentric video $\mathcal{V}^{\text{exo}} $ is used as condition and transformed into corresponding egocentric video $\mathcal{V}^{\text{ego}}$;
and (ii) \textit{Egocentric-to-exocentric translation}, where the egocentric video $\mathcal{V}^{\text{ego}}$ and reference image $\mathcal{I}^{\text{ref}}$ are used as the condition and transformed back into the exocentric video $\mathcal{V}^{\text{exo}}$.
Note that neither camera poses nor depth information is provided, which deliberately complicates the task to enforce robust generalization across different cameras and scenarios.
Importantly, our work addresses bidirectional ego-exo translation in dynamic wandering scenarios with freely moving cameras, which distinguishes it from previous unidirectional exo-to-ego methods~\cite{4Diff,ExoEgoV,Egoexo-gen} with static viewpoints.

\subsection{In-Context Perspective Alignment}
\label{sec:4.2}
Previous controllable video generation methods~\cite{CameraCtrl,MotionCtrl} typically train an auxiliary network to inject conditional information. Although this strategy is effective, it inevitably introduces extra parameters.
To address this limitation, we introduce an \textit{In-Context Perspective Alignment} paradigm, which efficiently performs in-context learning within a single backbone network, avoiding auxiliary networks.

\noindent
\textbf{Shared Latent Space.}
Recall the mentioned triplet sample $\langle \mathcal{V}^{\text{ego}}, \mathcal{V}^{\text{exo}}, \mathcal{I}^{\text{ref}} \rangle $, where $ \mathcal{V}^{\text{ego}} $ represents the egocentric video, $ \mathcal{V}^{\text{exo}} $ denotes the corresponding exocentric video, and $ \mathcal{I}^{\text{ref}} $ depicts the back-view character image.
During fine-tuning, instead of training an additional condition encoder, we utilize a unified VAE encoder $\mathcal{E}$ to encode them into a shared latent space $\langle z^{\text{ego}}_{0}, z^{\text{exo}}_{0}, z^{\text{ref}}_{0} \rangle$, as below:
\begin{equation}
    z^{\text{ego}}_{0} = \mathcal{E}(\mathcal{V}^{\text{ego}});\quad
    z^{\text{exo}}_{0} = \mathcal{E}(\mathcal{V}^{\text{exo}});\quad
    z^{\text{ref}}_{0} = \mathcal{E}(\mathcal{I}^{\text{ref}}),
\end{equation}
where all latent can share semantic space while eliminating additional encoders.

\noindent
\textbf{Different Noise Levels.}
Building upon the shared latent space, the target latent is noised in the same manner as the rectified flow~\cite{lipman2022flow}, while the conditional latent remains noise-free.
This strategy preserves the integrity of the conditional information while explicitly distinguishing the conditional latents from the target latent, enabling the model to learn their correspondence more effectively.

In detail, for exocentric-to-egocentric translation: the egocentric latent $z^{\text{ego}}_{0}$ is noised into $z^{\text{ego}}_{t}$ according to \cref{eq:1}, while the exocentric latent $z^{\text{exo}}_{0}$ remains noise-free; for egocentric-to-exocentric translation: the exocentric latent $z^{\text{exo}}_0$ is noised into $z^{\text{exo}}_{t}$ according to \cref{eq:1}, while the egocentric latent $z^{\text{ego}}_{0}$ and reference character latent $z^{\text{ref}}_{0}$ remains noise-free.
Note that the timestep of the noised latent is $t$, while that of the noise-free latent is $0$, accordingly.
In the following, we discuss how to enable efficient interaction between the conditional latent and the target latent in the egocentric-exocentric translation task.

\begin{figure}[t]
    \centering
    \includegraphics[width=0.98\linewidth]{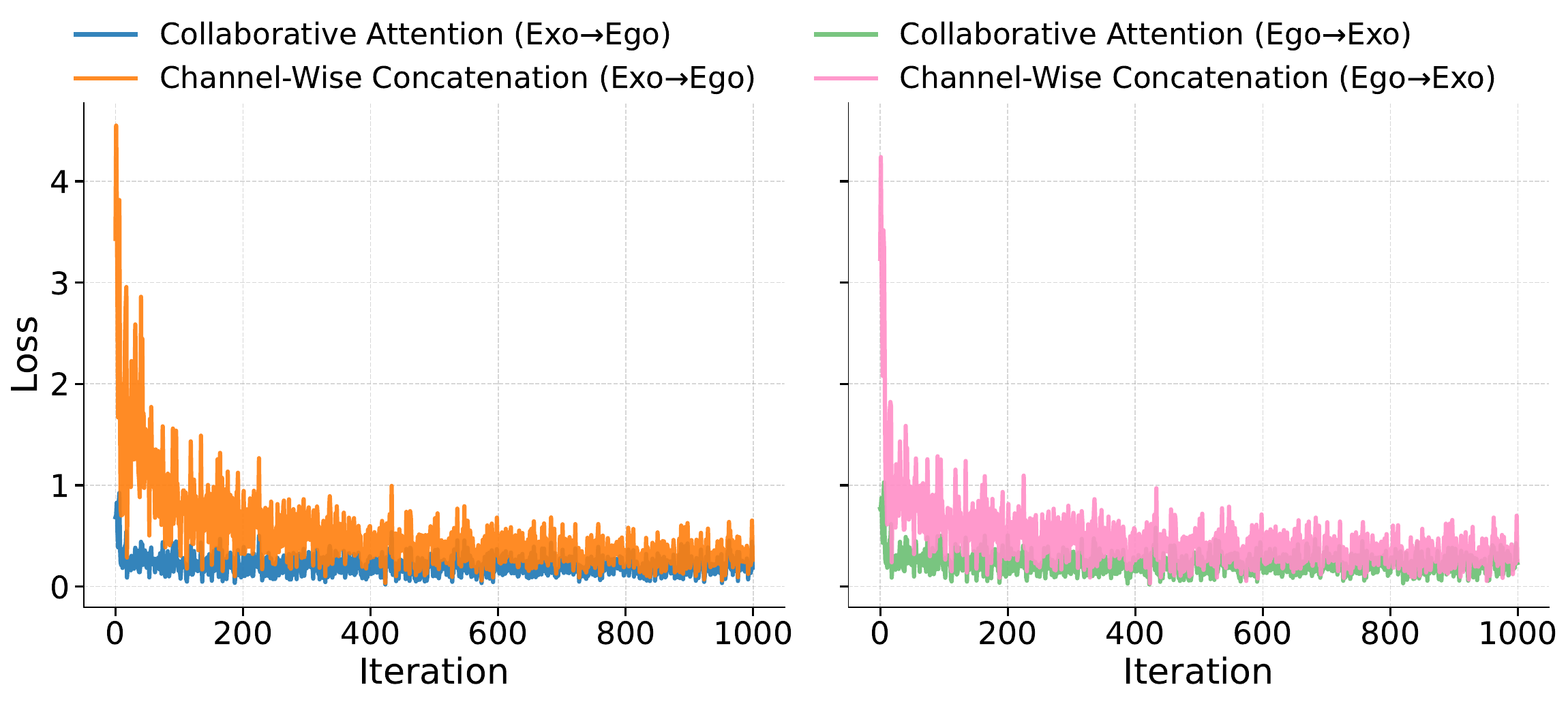}
    \caption{
    Loss comparison across different approaches on synthetic scenarios. \textit{Collaborative Attention} demonstrates faster convergence than \textit{Channel-Wise Concatenation}.
    }
    \label{fig:analysis}
\end{figure}

\noindent
\textbf{Collaborative Attention.}
To integrate the conditional latent into the target latent within a single backbone network, a simple approach is to concatenate them along the channel dimension for information fusion.
This only requires modifying input channels during the patchify stage, while keeping the rest of the backbone unchanged.
However, as shown in \cref{fig:analysis}, direct channel-wise concatenation leads to slow convergence during fine-tuning.
We attribute this to the channel fusion, which hinders explicit interaction between the two latent types.
To address this, we concatenate the conditional and target latents along the token dimension, integrating collaborative attention to enable explicit interaction.

Specifically, for exocentric-to-egocentric translation: we concatenate the noise-free exocentric latent $z^{\text{exo}}_{0}$  and the noisy egocentric latent $z^{\text{ego}}_{t}$ along the token dimension as the unified input $z_t^{\text{uni}}$;
for egocentric-to-exocentric translation: we concatenate the noise-free reference character latent $z^{\text{ref}}_{0}$, the noise-free exocentric latent $z^{\text{ego}}_{0}$, and the noised exocentric latent $z^{\text{exo}}_{t}$ along the token dimension as the unified input $z_t^{\text{uni}}$. Formally, this process can be formulated as follows:
\begin{equation}
    z^{\text{uni}}_{t} = 
    \begin{cases}
    \operatorname{TokenConcat}\left( [ z^{\text{exo}}_{0}, z^{\text{ego}}_{t} ] \right), & \text{exo}\!\to\!\text{ego} \\
    \operatorname{TokenConcat}\left( [ z^{\text{ref}}_{0}, z^{\text{ego}}_{0}, z^{\text{exo}}_{t} ] \right), & \text{ego}\!\to\!\text{exo}
    \end{cases}.
\end{equation}
During the self-attention process, the unified input $z_t^{\text{uni}}$ is projected through different learnable matrices $W_q$, $W_k$, and $W_v$, and then participates in multi-modal (conditional latent and target latent) attention interactions.
Formally, the attention output $\mathcal{O}$ can be formulated as follows:
\begin{equation}
    \mathcal{O} = \text{Softmax}\left(\frac{(z_t^{\text{uni}} \mathcal{W}_q)(z_t^{\text{uni}} \mathcal{W}_k)^\top}{\sqrt{d_k}}\right)(z_t^{\text{uni}} \mathcal{W}_v),
\end{equation}
where $d_k$ denotes the feature dimension.
In this way, the conditional and target latents share the same projection matrices for global interaction, thereby avoiding additional parameters from auxiliary networks.

As illustrated in \cref{fig:analysis}, the collaborative attention strategy enables the model to achieve faster convergence compared to simple channel-wise concatenation.

\begin{figure}[t]
    \centering
    \includegraphics[width=0.98\linewidth]{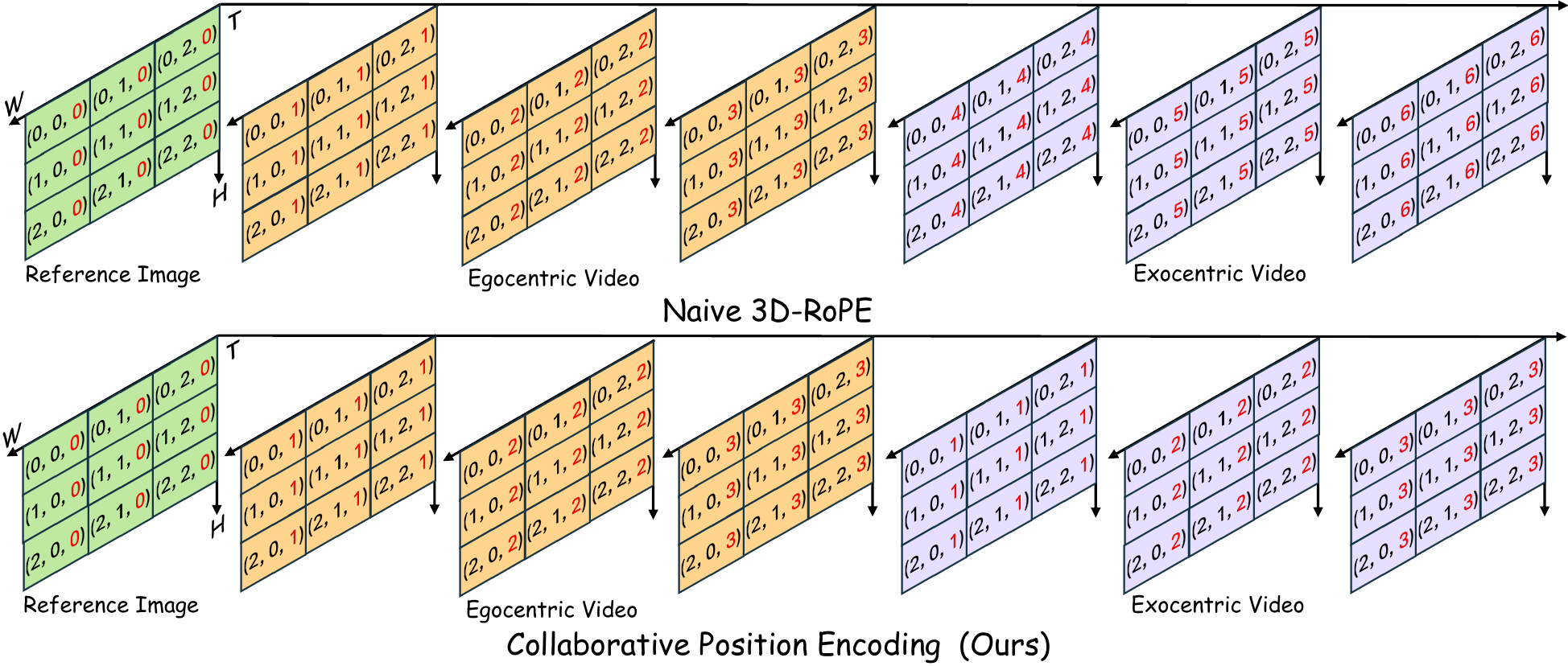}
    \caption{
    Toy examples between naive 3D RoPE and our collaborative position encoding.
    }
    \label{fig:examples}
\end{figure}

\subsection{Collaborative Position Encoding}
\label{sec:4.3}
To distinguish input tokens across spatial and temporal dimensions, rotary position embeddings are typically integrated into attention process according to \cref{eq:rope}.
Recall that our input $z^{\text{uni}}_t$ comprises both conditional and target latent. 
A straightforward approach is to treat them as a whole, uniformly encoding them along the temporal dimension.
However, as analyzed in our ablation study, this strategy often leads to incorrect cross-view correspondences between egocentric and exocentric videos.
In practice, the conditional and target latents exhibit a frame-wise alignment; uniformly encoding them disrupts this alignment, thereby hindering the learning such correspondences.
To overcome this limitation, we design a \textit{Collaborative Position Encoding} strategy to effectively learn the positional mapping between the two latent types, with the toy examples show in \cref{fig:examples}.

In detail, the unified latent $z^{\text{uni}}$ is first decomposed into corresponding conditional and target latents, which are then independently encoded to share the same position embeddings along the temporal dimension. This strategy enforces a frame-wise positional correspondence.
Formally, the position embeddings $\mathcal{P}^{\text{uni}}$ for two sub-tasks are formulated as follows:
\begin{equation}
\mathcal{P}^{\text{uni}}=
    \begin{cases}
    \operatorname{Concat}\!\big([R(z^{\text{exo}}_{0}), R(z^{\text{ego}}_{t})\big]), & \text{exo}\!\to\!\text{ego} \\
    \operatorname{Concat}\!\big([R(z^{\text{ref}}_{0}), R(z^{\text{ego}}_{0}), R(z^{\text{exo}}_{t})\big]), & \text{ego}\!\to\!\text{exo}
    \end{cases},
\end{equation}
where $R$ denotes the positional encoding operation.
Subsequently, the collaborative position embeddings $\mathcal{P}^{\text{uni}}$ are integrated into the attention process according to \cref{eq:rope}, thus facilitating the interaction between different latents.

\subsection{Efficient Fine-Tuning via LoRA}
\label{sec:4.4}
In this work, we adopt the video generation model Wan2.2-5B~\cite{wan2025wan} as our backbone.
Given the pretrained model encodes strong priors from large-scale video data, we fine-tune it with LoRA~\cite{hu2022lora} to efficiently preserve these priors under limited resources.
In detail, LoRA injects trainable low-rank matrices $\mathcal{A} \in \mathbb{R}^{r \times k}$ and $\mathcal{B} \in \mathbb{R}^{d \times r}$ into specific layers of the model, while keeping the corresponding original weights $\mathcal{W} \in \mathbb{R}^{d \times k}$ frozen to preserve the priors.
Here, $r$ is the rank, $k$ is the input dimension, and $d$ is the output dimension, with $r \ll \min(d, k)$.
Once integrated with LoRA, the forward computation is modified below:
\begin{equation}
    y' = y + \Delta y = \mathcal{W}  \cdot x + \mathcal{B} \cdot \mathcal{A} \cdot x,
\end{equation}
where $x$ is the input to specific layers and $y'$ is the corresponding output. By default, matrix $\mathcal{B}$ is initialized to zeros to stabilize the fine-tuning process.

During fine-tuning, the backbone is integrated with the proposed \textit{In-Context Perspective Alignment} paradigm and \textit{Collaborative Position Encoding} strategy, and optimized using the flow-matching loss defined in \cref{eq:2} as supervision.

\begin{figure}[t]
    \centering
    \includegraphics[width=0.98\linewidth]{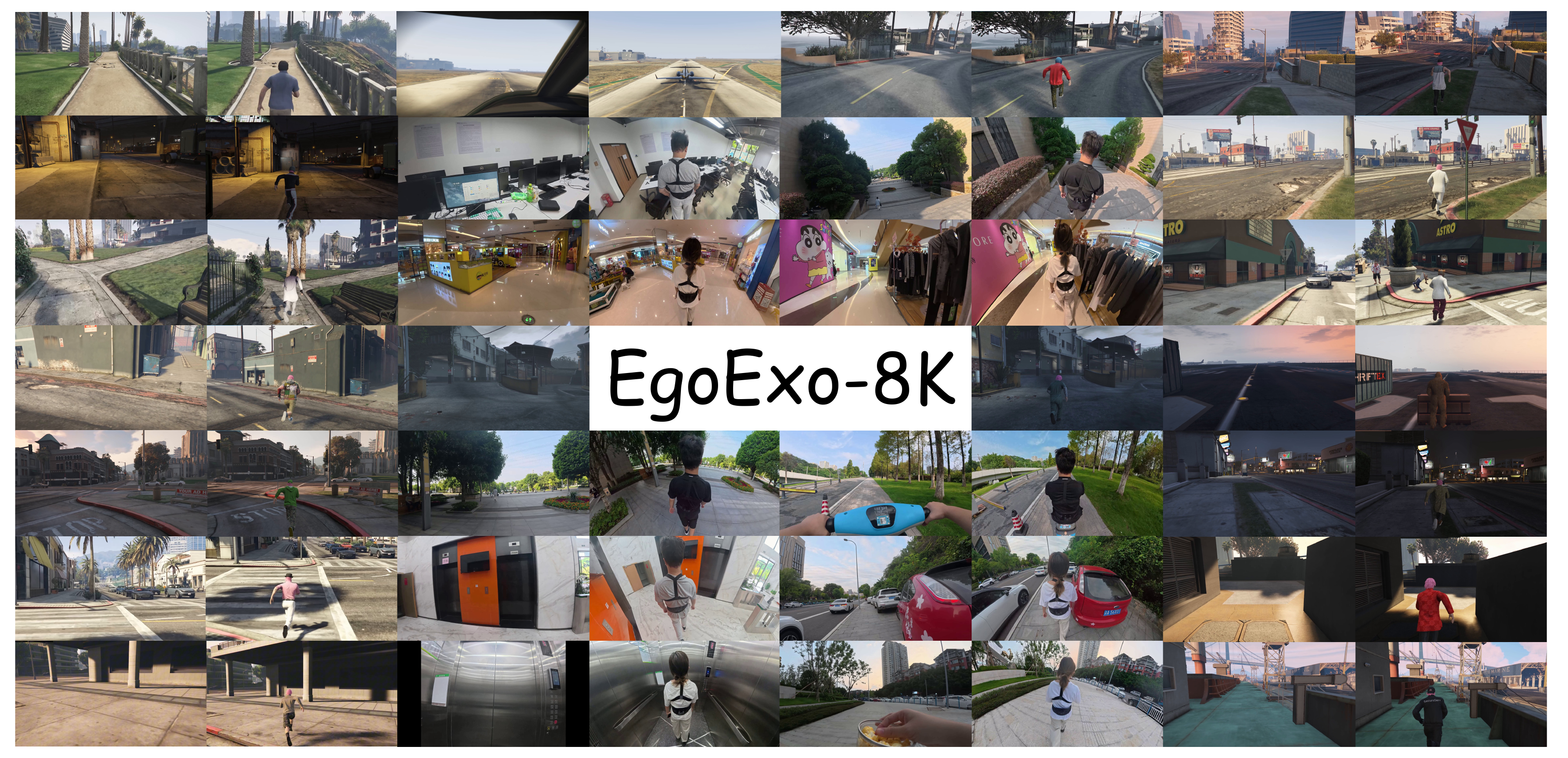}
    \caption{
    Representative samples of our carefully curated EgoExo-8K dataset. Compared to prior egocentric-exocentric datasets~\cite{EgoExo4D,H2O,TACO}, EgoExo-8K uniquely supports dynamic scene navigation and flexible camera tracking, encompassing both synthetic and real-world scenarios across diverse indoor and outdoor environments.
    }
    \label{fig:datasets}
\end{figure}

\subsection{EgoExo-8K}
\label{sec:4.5}
High-quality egocentric-exocentric data is a key prerequisite for models to learn cross-view correspondences.
Although prior works~\cite{EgoExo4D,H2O,TACO} have introduced paired egocentric–exocentric datasets, these datasets are collected from static viewpoints in confined scenarios, involving limited camera dynamics.
Such static setups contradict our core vision: modeling a \textit{``WorldWander''} paradigm, where a character freely wanders and navigate across dynamic environments and the viewpoint can seamlessly switch between egocentric and exocentric perspectives.
To bridge this gap, we curate the \textit{EgoExo-8K} dataset (see \cref{fig:datasets}), a dynamic and scene-rich collection of synchronized egocentric-exocentric triplets for perspective transformation during open-world wandering. 
Each triplet sample is organized as $\langle \mathcal{V}^{\text{ego}}, \mathcal{V}^{\text{exo}}, \mathcal{I}^{\text{ref}} \rangle$, where $\mathcal{V}^{\text{ego}}$ and $\mathcal{V}^{\text{exo}}$ denote the egocentric and exocentric videos, while $\mathcal{I}^{\text{ref}}$ is the corresponding reference image.
In summary, EgoExo-8K contains approximately 4,000 triplets derived from both synthetic and real-world scenarios.
More samples and analyses can be found in the \textbf{Appendix}.

\noindent
\textbf{Synthetic Scenarios.}
The synthetic subset is recorded within the GTA-5 environment, offering a photorealistic and controllable virtual world.
In detail, we adopt a two-player collection protocol: one player controls the selected character and wanders across scenes from an egocentric perspective, while the other captures the exocentric perspective via \emph{spectator mode}.
Both recordings are strictly synchronized to ensure temporal consistency, with the selected character saved as the reference image.
In total, we construct 400+ paired video sequences along with their corresponding reference images, where each video lasts more than one minute. These samples are subsequently filtered and segmented into 5-second clips (300 frames), yielding exactly 4,000 triplet samples.
These triplet samples cover diverse indoor and outdoor scenes, featuring navigation, camera tracking, and dynamic interactions such as manipulation and task execution.

\noindent
\textbf{Real-World Scenarios.}
The real-world subset is captured across everyday environments, providing a natural and complex physical counterpart.
Specifically, we adopt a dual-camera collection protocol: a participant equipped with a dual-camera setup wanders across diverse scenes, where a head-mounted action camera (Ac-tion5Pro) captures the egocentric perspective, while a rear-mounted panoramic camera (Osmo360) simultaneously records the exocentric view.
Similarly, both recordings are strictly synchronized to ensure temporal consistency, while the reference image is randomly sampled from the exocentric video.
In total, we construct 10h+ paired video sequences along with their corresponding reference images. These samples also subsequently filtered and segmented into 5-second clips (300 frames), yielding 4,000 triplet samples.
These triplet samples also span indoor and outdoor scenes that capture human activities (\emph{e.g.}, walking, cleaning, and carrying objects), providing realistic cross-view correspondences.

\section{Experiments}
\subsection{Experimental details.}

\noindent
\textbf{Implementation Details.}
Our \method is built upon the state-of-the-art video diffusion transformer Wan2.2-5B~\cite{wan2025wan}, into which LoRA~\cite{hu2022lora} modules are integrated within the attention layers, with the rank $r=80$. 
Each video clip is resized and cropped to $704 \times 1280$ to match the original configuration of Wan2.2-5B, with the number of frames $F=49$.
Considering the inherent differences between the egocentric-to-exocentric and exocentric-to-egocentric tasks, we fine-tune the model for each task using the same configuration.
In detail, we use a batch size of $4$ per GPU and the AdamW optimizer (learning rate $1\times10^{-4}$, weight decay $1\times10^{-2}$). All experiments are conducted on four NVIDIA H200 GPUs. Training takes about two days, while inference requires about three minutes.

\noindent
\textbf{Evaluation Settings.}
Our task aims to tackle mutual egocentric-exocentric perspective translation.
However, direct comparisons with existing exo-to-ego methods are infeasible: the implementations of 4Diff~\cite{4Diff} and EgoExoGen~\cite{Egoexo-gen} are not publicly available, while EgoExo-V~\cite{ExoEgoV} requires four exocentric video inputs, creating a fundamental mismatch with our setting.
In light of these, we alternatively select two categories of closely related methods to serve as baselines:
(i) \textit{video-to-video translation}, AnyV2V~\cite{ku2024anyv2v} and I2VEdit~\cite{ouyang2024i2vedit};
(ii) \textit{perspective re-orientation}, TrajectoryCrafter~\cite{TrajectoryCrafter} and RecamMaster~\cite{Recammaster}.
Implementation details for all baselines are provided in the \textbf{Appendix}.
We evaluate the effectiveness of these approaches using our curated EgoExo-8K dataset.
Considering the domain gap between synthetic and real-world scenes, we fine-tune the model individually for each scene and reserve 50 corresponding triplets as the test set for evaluation.

\begin{table*}[!t]
\centering
\caption{
Automatic evaluation across methods on both the exocentric-egocentric translation tasks.
We report results for both synthetic and real-world scenarios, with the best result in \textbf{bold} and the second-best result \underline{underlined}. Note that in the zero-shot setting, TrajectoryCrafter is unable to generate customized characters; therefore, \textit{we do not report its CLIP-I score}.
$^{*}$ indicates fine-tuning using our egoexo-8K datasets.
}
\resizebox{\textwidth}{!}{
\begin{tabular}{l cccc ccccc}
\toprule
\multirow{2}{*}{Methods} 
& \multicolumn{4}{c}{Exocentric-to-Egocentric} 
& \multicolumn{5}{c}{Egocentric-to-Exocentric} \\
\cmidrule(lr){2-5} \cmidrule(lr){6-10}
& LPIPS$\downarrow$ & SSIM$\uparrow$ & FVD$\downarrow$ & VBench$\uparrow$ 
& LPIPS$\downarrow$ & SSIM$\uparrow$ & FVD$\downarrow$ & VBench$\uparrow$ & CLIP-I$\uparrow$ \\
\midrule
\multicolumn{10}{c}{\textit{Synthetic Scenarios}} \\
AnyV2V~\cite{ku2024anyv2v}                & 0.6627 & 0.4870 & 990.3 & 0.7595 & 0.6809 & 0.4516 & 1285.9 & 0.7311 & 0.6240 \\
I2VEdit$^{*}$~\cite{ouyang2024i2vedit}          & 0.6272 & 0.5154 & 477.2 & 0.8037 & 0.6385 & 0.4940 & 737.4 & 0.7880 & \underline{0.7470} \\
TracjtoryCrafter~\cite{TrajectoryCrafter}  & 0.6109 & 0.5068 & 743.2 & 0.7967 & 0.6112 & 0.5031 & 745.4 & \underline{0.8171} & – \\
ReCamMaster~\cite{Recammaster}             & 0.6516 & 0.4039 & 738.7 & 0.7634 & 0.6178 & 0.4629 & 915.7 & 0.7780 & 0.6537 \\
ReCamMaster$^{*}$~\cite{Recammaster}             & \underline{0.6039} & \underline{0.5365} & \underline{298.9} & \underline{0.8098} & \underline{0.5732} & \underline{0.5125} & \underline{531.5} & 0.8116 & 0.7129 \\
\rowcolor{gray!25}
\method(Ours)                             & \textbf{0.5811} & \textbf{0.5416} & \textbf{230.3} & \textbf{0.8110} & \textbf{0.5550} & \textbf{0.5215} & \textbf{353.7} & \textbf{0.8395} & \textbf{0.7672} \\
\midrule
\multicolumn{10}{c}{\textit{Real-World Scenarios}} \\
AnyV2V~\cite{ku2024anyv2v}                & 0.7227 & 0.3394 & 1789.3 & 0.6813 & 0.7182 & 0.2885 & 1451.2 & 0.6705 & 0.5836 \\
I2VEdit$^{*}$~\cite{ouyang2024i2vedit}          & 0.6945 & 0.3428 & 1226.3 & \underline{0.7996} & 0.6892 & \underline{0.3253} & \underline{904.4} & \underline{0.7815} & 0.6601 \\
TracjtoryCrafter~\cite{TrajectoryCrafter}  & 0.7058 & 0.3127 & 1238.4 & 0.7977 & 0.6928 & 0.3215 & 1126.5 & 0.7583 & – \\
ReCamMaster$^{*}$~\cite{Recammaster}             & \underline{0.6752} & \underline{0.3476} & \underline{716.1} & 0.7933 & \underline{0.6699} & 0.3048 & 968.1 & 0.7787 & \underline{0.7020} \\
\rowcolor{gray!25}
\method(Ours)                             & \textbf{0.6054} & \textbf{0.3567} & \textbf{416.5} & \textbf{0.8333} & \textbf{0.5357} & \textbf{0.3469} & \textbf{374.7} & \textbf{0.8142} & \textbf{0.7214} \\
\bottomrule
\end{tabular}
}
\label{tab:quantitative}
\end{table*}

\subsection{Quantitative Results}
In this section, we conduct quantitative comparisons across baselines on both the exocentric-to-egocentric and egocentric-to-exocentric video translation tasks.
The evaluation focuses on both synthetic and real-world scenarios, analyzed from two complementary perspectives: \textit{Automatic Evaluation} that provides objective comparison, and \textit{User Study} that captures subjective comparison.

\noindent
\textbf{Automatic Evaluation.}
We evaluate the generated videos across different tasks and from multiple dimensions.
For the \textit{exocentric-to-egocentric} translation: LPIPS and SSIM between generated and ground-truth videos are used to evaluate perspective alignment, FVD~\cite{unterthiner2018towards} with respect to the corresponding ground-truth is used to assess distribution alignment, and the average VBench~\cite{huang2024vbench} score is used to measure overall generation quality.
For the \textit{egocentric-to-exocentric} translation: we report the same set of metrics and additionally include CLIP-I to evaluate the semantic consistency between the characters in the generated videos and the reference images.
We report results both on synthetic and real-world scenarios, with the details illustrated in \cref{tab:quantitative}. Note that in the zero-shot setting, TrajectoryCrafter~\cite{TrajectoryCrafter} is unable to generate customized characters; therefore, we do not report its CLIP-I score.
In synthetic scenarios, our \method outperforms all baselines in terms of distribution alignment, perspective alignment, overall quality, and character consistency, across both directions of egocentric–exocentric video translation.
With respect to real-world scenarios, \method also outperforms all baselines across all evaluation dimensions, showing superior performance in the egocentric-exocentric video translation tasks.
An additional point worth noting is that we compare the performance of RecamMaster with and without fine-tuning on our EgoExo-8K; the fine-tuned version shows superior performance, further validating the effectiveness of our datasets.

\begin{figure}[t]
    \centering
    \includegraphics[width=0.98\linewidth]{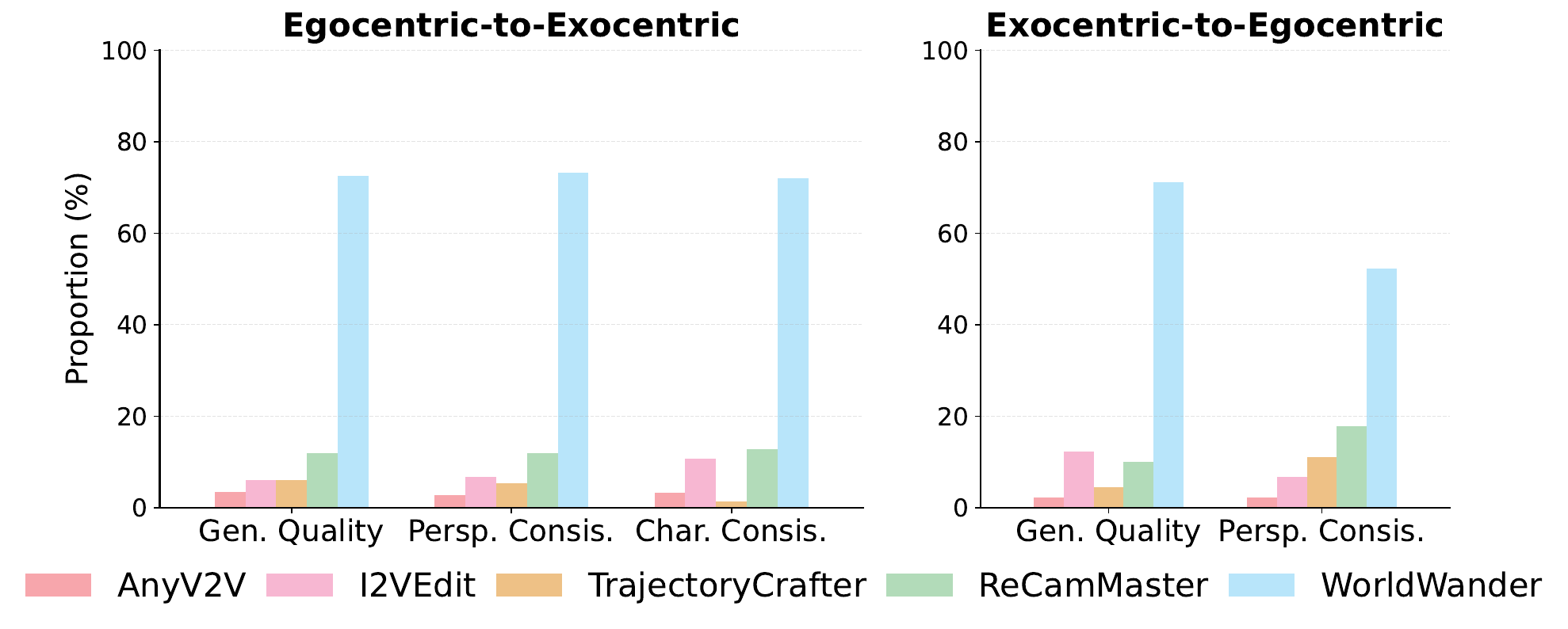}
    \caption{
    User study across baselines on the exocentric-egocentric video translation tasks. We report average results of the synthetic and real-world scenarios.
    }
    \label{fig:user_study}
\end{figure}

\noindent
\textbf{User Study.}
To further evaluate different methods on bidirectional \textit{egocentric–exocentric} video translations, we conduct a user study to assess human perceptual preferences.
In detail, we design a questionnaire-based evaluation covering both real and synthetic scenes.
For exocentric-to-egocentric translation, participants are asked to select the best output among all methods based on (i) generation quality and (ii) perspective consistency.
For egocentric-to-exocentric translation, participants are asked to select the best output using the same criteria and additionally evaluate (iii) character consistency.
In total, we collect 630 valid responses and report the averaged results for both synthetic and real-world scenarios, with the details illustrated in \cref{fig:user_study}.
Clearly, our \method outperforms all baselines across different subjective metrics, further demonstrating its human preferences in the egocentric-exocentric video translation task.

\begin{figure*}[t]
    \centering
    \includegraphics[width=0.98\linewidth]{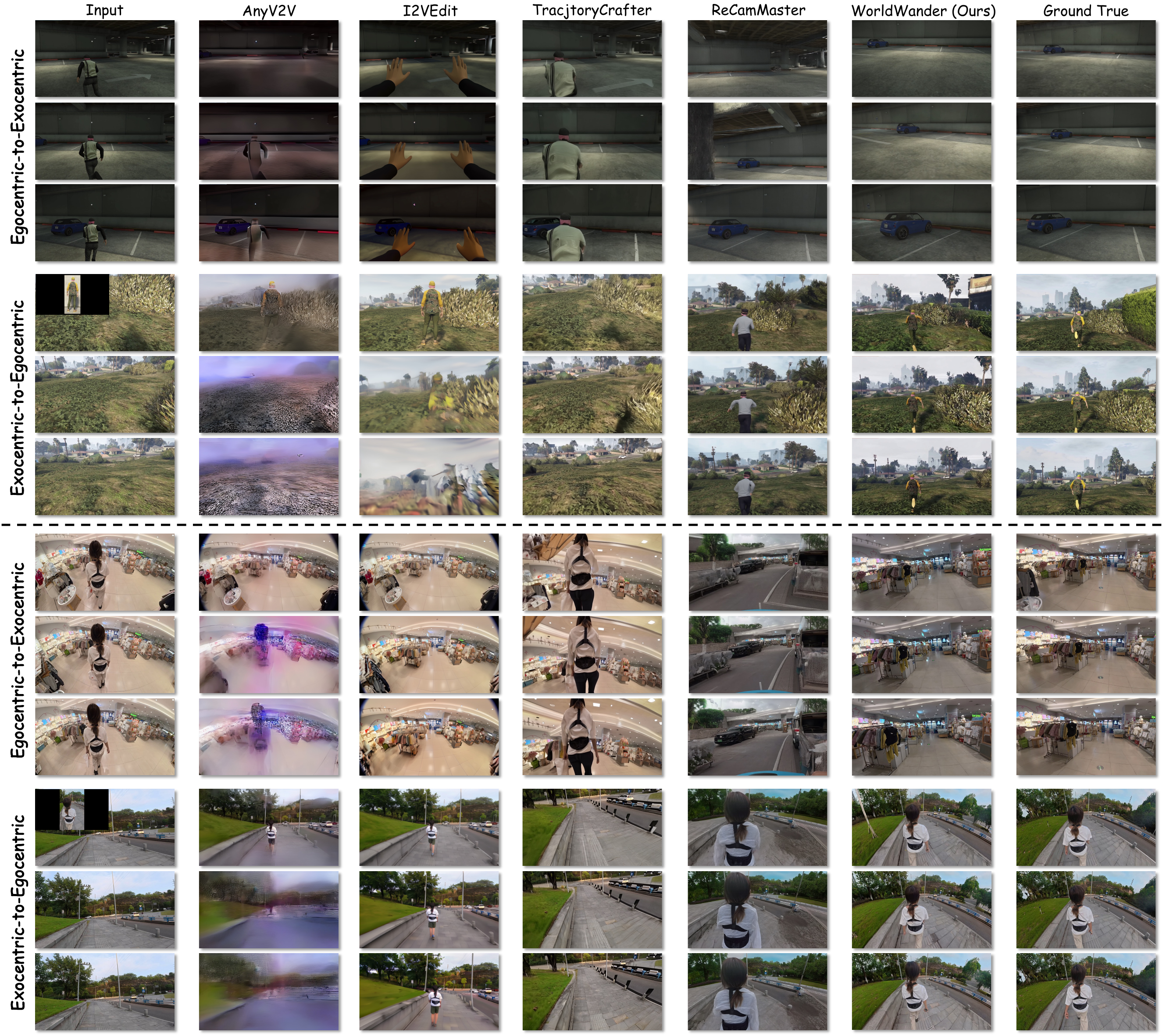}
    \caption{
    Qualitative comparison of different methods on both exocentric-to-egocentric and egocentric-to-exocentric video translation tasks. Best viewed with zooming in.
    }
    \label{fig:qualitative}
\end{figure*}

\subsection{Qualitative Results}
We conduct a detailed comparison across baselines for both exocentric-to-egocentric and egocentric-to-exocentric translations, and visualize the results for synthetic and real-world scenarios in \cref{fig:qualitative}.
In synthetic scenarios, traditional video-to-video translation methods like AnyV2V~\cite{ku2024anyv2v} and I2VEdit~\cite{ouyang2024i2vedit} fail to model perspective correspondences.
Recent perspective re-orientation approaches, such as Trajectory~\cite{TrajectoryCrafter}, which only support perspective changes, struggle to preserve identity consistency and handle identity removal in exocentric-to-egocentric transformations.
In contrast, our \method exhibits higher perspective consistency and character consistency than other methods, while maintaining superior generation quality.
In real-world scenarios, \method also exhibits superior synchronization and quality, further highlighting its generalization.
In addition to viewpoint translation in scene wandering, we also support translation for specific human-object interactions, and we visualize representative examples in \cref{fig:HOI}.
More comparisons and visualization are provided in \textbf{Appendix}.

\begin{figure*}[t]
    \centering
    \includegraphics[width=0.98\linewidth]{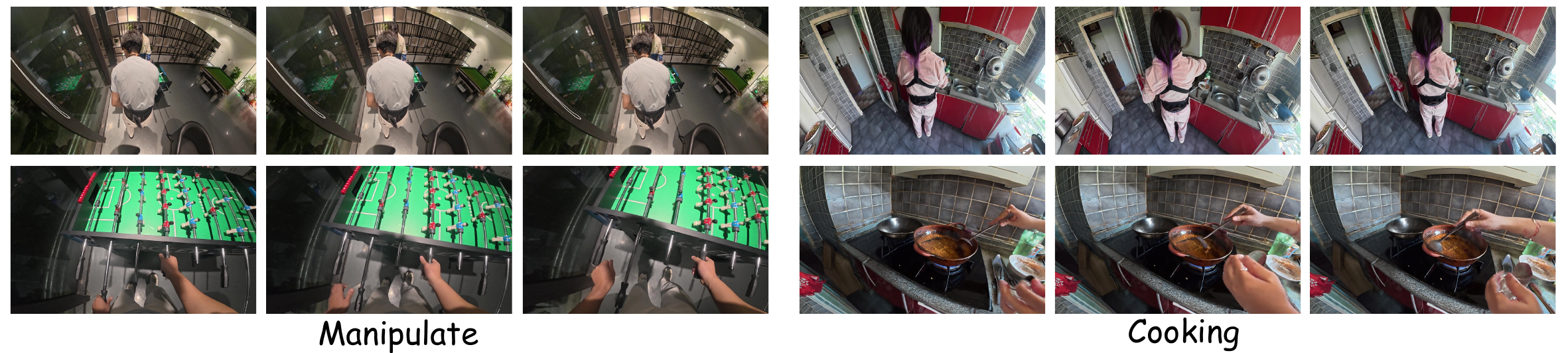}
    \caption{
    Representative examples for specific human-object interaction.
    }
    \label{fig:HOI}
\end{figure*}

\subsection{Ablation Studies}
In this section, we study the effectiveness of the core components: \textit{In-Context Perspective Alignment} and \textit{Collaborative Position Encoding}, and corresponding LoRA rank.
Due to limited resources, we study only on synthetic scenarios.

\begin{table*}[!t]
\centering
\caption{Quantitative ablation about two components: \textit{In-Context Perspective Alignment} and \textit{Collaborative Position Encoding}.
We compare two variants: \textit{channel concatenation} and \textit{uniform position encoding}, on both the exocentric-to-egocentric and egocentric-to-exocentric translation task.
Due to limited resources, we only report the ablation results for synthetic scenarios, with the best result in \textbf{bold}.
}
\resizebox{\textwidth}{!}{
\begin{tabular}{l cccc ccccc}
\toprule
\multirow{2}{*}{Variants} 
& \multicolumn{4}{c}{Exocentric-to-Egocentric} 
& \multicolumn{5}{c}{Egocentric-to-Exocentric} \\
\cmidrule(lr){2-5} \cmidrule(lr){6-10}
& LPIPS$\downarrow$ & SSIM$\uparrow$ & FVD$\downarrow$ & VBench$\uparrow$ 
& LPIPS$\downarrow$ & SSIM$\uparrow$ & FVD$\downarrow$ & VBench$\uparrow$ & CLIP-I$\uparrow$ \\
\midrule
Channel Concat.            & 0.6008 & 0.5221 & 280.2 & 0.8072 & 0.6205 & 0.4708 & 524.3 & 0.8377 & 0.7310 \\
Uni. Timestep              & 0.5963 & 0.5209 & 287.7 & 0.8063 & 0.5551 & 0.5211 & 365.9 & 0.8389 & 0.7429 \\
Uni. Pos. Enc.             & 0.5907 & 0.5325 & 280.3 & 0.8044 & 0.5771 & 0.5083 & 356.5 & 0.8357& 0.7562 \\
\midrule
Full Comp.($r$=64)         & 0.5931 & 0.5260 & 282.6 & 0.8083 & \textbf{0.5471} & 0.5165 & 362.9 & 0.8306 & 0.7461 \\
\rowcolor{gray!25}
Full Comp.($r$=80)         & \textbf{0.5811} & \textbf{0.5416} & \textbf{230.3} & 0.8110 & 0.5550 & \textbf{0.5215} & 353.7 & \textbf{0.8395} & \textbf{0.7672} \\
Full Comp.($r$=128)        & 0.5819 & 0.5351 & 273.0 & \textbf{0.8122} & 0.5587 & 0.5202 & \textbf{344.9} & 0.8392 & 0.7556 \\
\bottomrule
\end{tabular}
}
\label{tab:ablation}
\end{table*}

\begin{figure*}[t]
    \centering
    \includegraphics[width=0.98\linewidth]{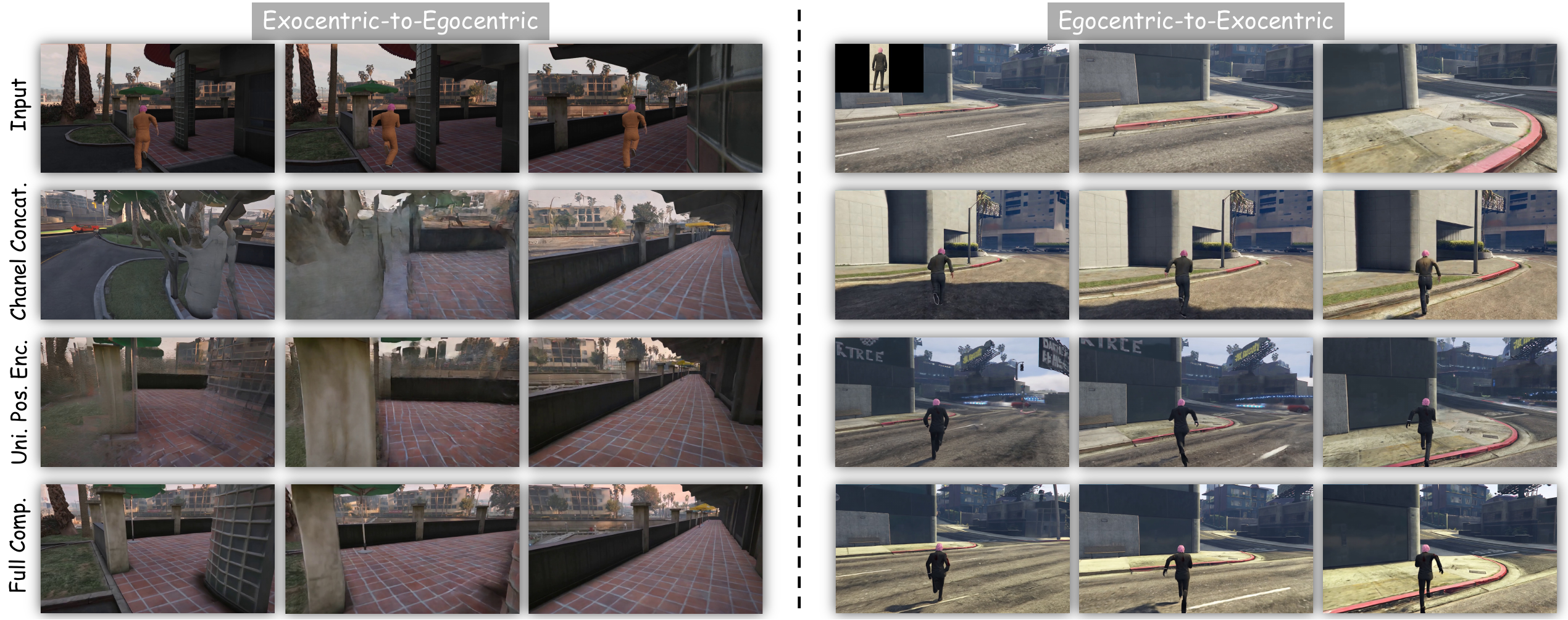}
    \caption{
    Qualitative ablation about two components: \textit{In-Context Perspective Alignment} and \textit{Collaborative Position Encoding}. Best viewed with zooming in.
    }
    \label{fig:ablation}
\end{figure*}

\noindent
\textbf{In-Context Perspective Alignment.}
The core of in-context perspective alignment is the collaborative attention mechanism, which enables the clean conditional latent and the noisy target latent to perform in-context learning.
To validate the effectiveness of this, we compare it with a variant that concatenates the two latents along the channel dimension.
In \cref{tab:ablation}, collaborative attention consistently outperforms channel-wise fusion across all metrics in egocentric-exocentric video translation.
The qualitative results in \cref{fig:ablation} also support this conclusion.
Additionally, we compare a uniform timestep with varying timestep; the results in \cref{tab:ablation} also demonstrate the superiority of our design.

\noindent
\textbf{Collaborative Position Encoding.}
Recall that our collaborative positional encoding independently encodes the conditional and target latents, thereby sharing the same embeddings along the temporal dimension and achieving better frame-level perspective correspondence.
To further evaluate its superiority, we compare it with a variant that uniformly encodes the conditional and target latents, where the positional embeddings differ across all frames.
The quantitative results are reported in \cref{tab:ablation}, while the qualitative results are illustrated in \cref{fig:ablation}.
Our collaborative position encoding strategy consistently outperforms the uniform position encoding in egocentric-exocentric video translation.

\noindent
\textbf{LoRA Rank.}
We compare the performance of LoRA with ranks of $64$, $80$, and $128$, with the results shown in \cref{fig:ablation}.
Under the limited datas, a rank of $80$ shows superior overall performance, so we ultimately selecte it as final setting.

\section{Conclusion}
This paper formulates the novel \textit{egocentric-exocentric video perspective translation} task.
To achieve this, we propose WorldWander, an in-context learning framework tailored for translating between egocentric and exocentric worlds in video generation.
Leveraging advanced video diffusion transformers, \method performs LoRA fine-tuning combined with the proposed \textit{In-Context Perspective Alignment} and \textit{Collaborative Position Encoding}. These strategies enable effective modeling of cross-view correspondences and character consistency.
To facilitate this research, we further curate \textit{EgoExo-8K}, a dynamic and scene-rich dataset featuring synchronized egocentric–exocentric triplet samples collected from both synthetic and real-world environments.
Extensive experiments validate the effectiveness of \method, demonstrating superior perspective synchronization, character consistency, and generalization across diverse scenarios.

\section*{Acknowledgements}
This project is supported by the Ministry of Education, Singapore, under the Academic Research Fund Tier 1 (FY2023).

\clearpage
\appendix

\section{Implementation Details about Baselines}
Recall that in the main experiments, we compare \method with four representative baselines: AnyV2V~\cite{ku2024anyv2v}, I2VEdit~\cite{ouyang2024i2vedit}, TrajectoryCrafter~\cite{TrajectoryCrafter}, and RecamMaster~\cite{Recammaster}.
To ensure a fair comparison, all methods are first applied at their optimal resolutions and then resized to $704 \times 1024$, with a fixed frame count of $F=49$.
The reference image is center-padded with black borders to match target resolution.
We outline the implementation details below:

AnyV2V is a training-free video-to-video translation method that takes both the edited first frame and the original video as inputs.
To adapt it to our gocentric-to-exocentric task, we use the state-of-the-art Qwen-image-edit~\cite{wu2025qwen} to edit the first frame, transforming it into the corresponding first-person or third-person perspective.
Both the edited first frame and the specific-perspective video are then used as inputs to generate the alternative perspective video.

I2VEdit is a one-shot tuning video-to-video translation approach that also takes both the edited first frame and the original video as inputs.
To adapt it to our task, we also employ the Qwen-image-edit~\cite{wu2025qwen} to edit the first frame, transforming it into the target perspective.
Both the edited first frame and the specific-perspective video are then used for fine-tuning the model, which subsequently generates the alternative perspective video.

TrajectoryCrafter is a perspective re-orientation method that takes an original video and predefined camera parameters to generate a video from the corresponding perspective.
Unfortunately, the official repository does not provide fine-tuning code. 
To adapt it to our gocentric-to-exocentric task, we use its pretrained weights to perform inference in a zero-shot setting.
In detail, we collect the camera poses for converting between egocentric and exocentric  perspective videos in the training set, and use them directly as model inputs to generate results on the test set.

RecamMaster is also a perspective re-orientation method that takes an original video and predefined camera parameters to generate a video from the corresponding perspective. To adapt it to our gocentric-to-exocentric task, we fine-tuned the model using the official code on our curated EgoExo-8K dataset, and then performed inference using the fine-tuned weights.
Similar to \method, we disregard the dependency on camera poses and treat the task as a video-to-video translation to ensure fair comparisons.

\begin{figure*}[!t]
    \centering
    \includegraphics[width=0.98\textwidth]{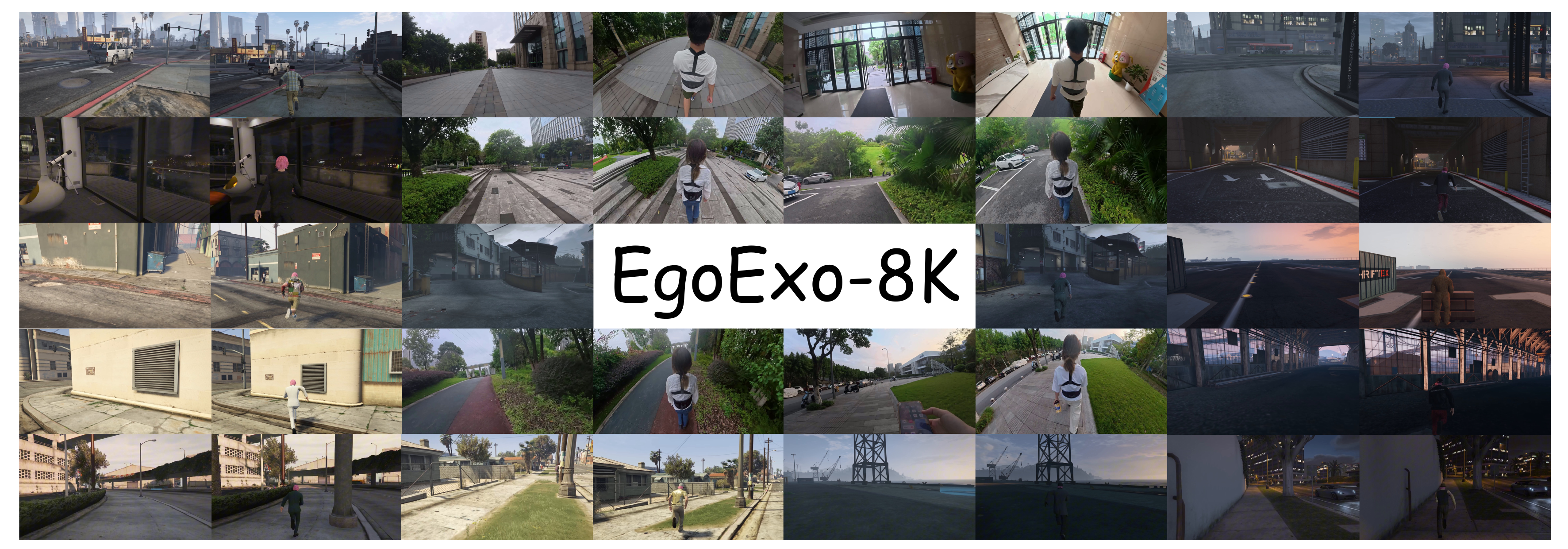}
    \caption{
    Additional examples of our curated EgoExo-8K. It features diverse synthetic and real-world indoor and outdoor scenarios. Best viewed with zooming in.
    }
    \label{fig:datasets_full}
\end{figure*}

\begin{figure*}[!t]
    \centering
    \includegraphics[width=0.98\textwidth]{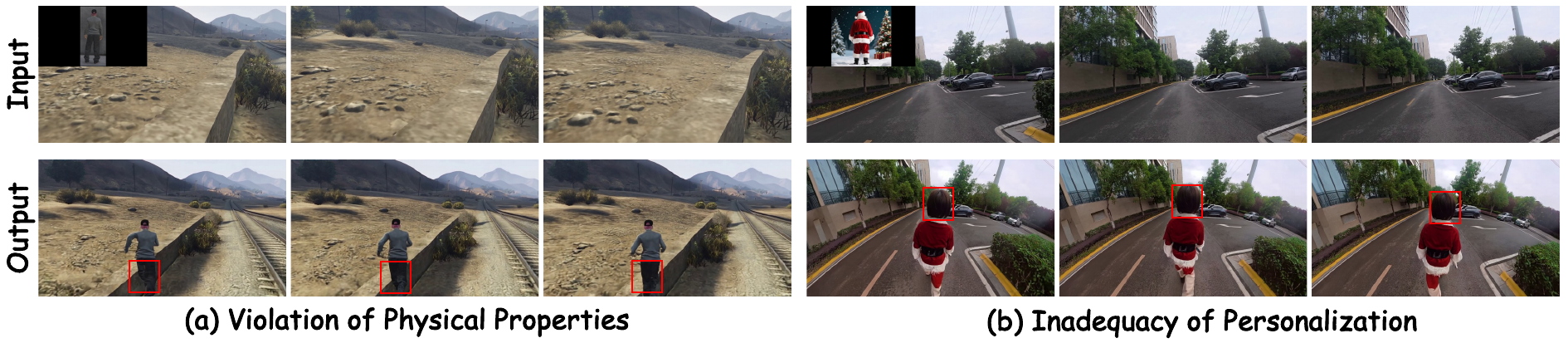}
    \caption{
    Failure cases of our \method. It includes (a) \textit{Violation of Physical Properties} and (b) \textit{Inadequacy of Personalization. Best viewed with zooming in.}
    }
    \label{fig:limitation}
\end{figure*}

\section{Details  and Examples about EgoExo-8K}
Our EgoExo-8K dataset includes both synthetic and real-world scenarios, covering diverse environments such as deserts, fields, urban areas, and various indoor and outdoor settings. We present additional paired examples in \cref{fig:datasets_full}. During fine-tuning, these paired videos are segmented into 5-second clips (300 frames each). We describe the data collection process for these scenes in detail below:

\begin{figure*}[!t]
    \centering
    \includegraphics[width=0.98\textwidth]{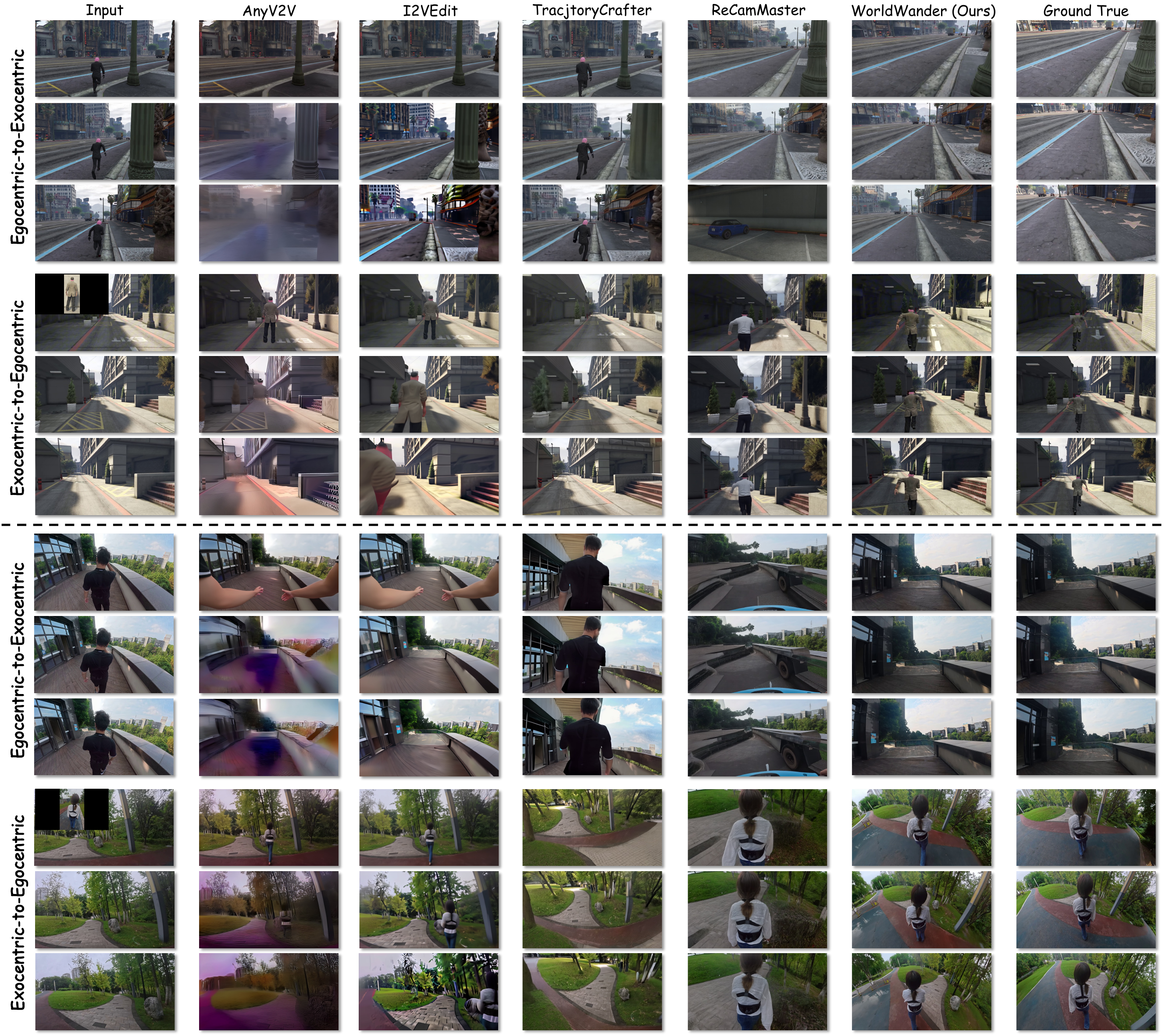}
    \caption{
    Additional qualitative comparison across approaches on the exocentric-egocentric video translations task. Best viewed with zooming in.
    }
    \label{fig:additional_qualitative}
\end{figure*}

\begin{figure*}[!t]
    \centering
    \includegraphics[width=0.98\textwidth]{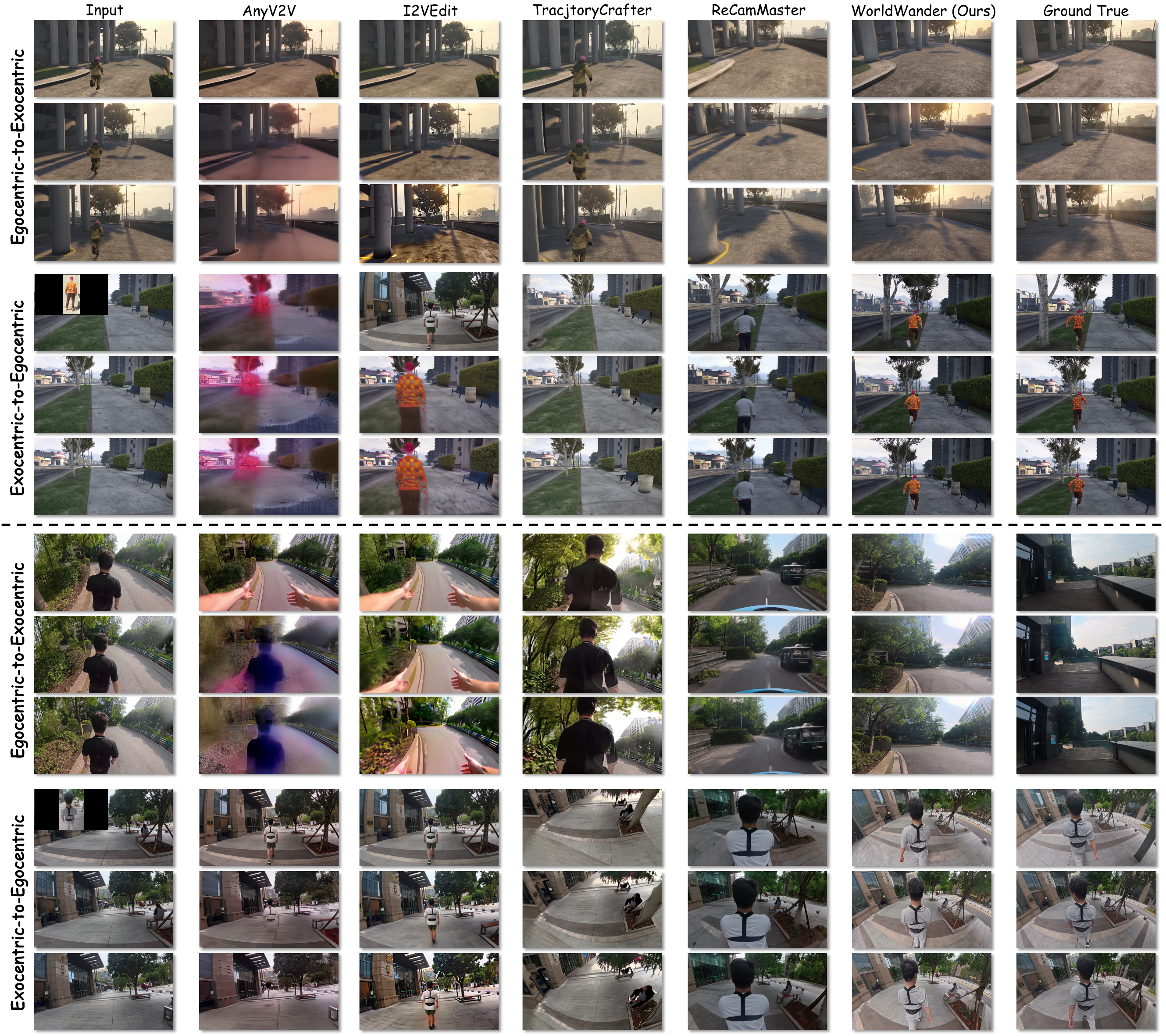}
    \caption{
    Additional qualitative comparison across approaches on the exocentric-egocentric video translations task. Best viewed with zooming in.
    }
    \label{fig:additional_qualitative2}
\end{figure*}

\textbf{Synthetic Scenarios.}
We collect paired first-person and third-person videos from GTA-5 using the two-player data collection protocol described in the main paper. Both perspectives have a resolution of $1440 \times 2560$ and a frame rate of $60$ fps.
For each scenario, we change the character's outfit and capture about one minute of traversal footage. Before recording, the outfitted character is photographed as a reference image.
To eliminate text information at the edges of the recorded game interface, we resize the longer side to $1505$ while maintaining the aspect ratio, then apply a center crop to $704 \times 1280$ (the resolution used in wan2.2~\cite{wan2025wan}). This ensures the final video frames are free of extraneous text.

\textbf{Real-World Scenarios.}
We capture synchronized first-person and third-person videos in real-world environments using the dual-camera collection protocol described in the main paper. Both perspectives have a resolution of $1080 \times 1920$ and a frame rate of $60$ fps.
As with the synthetic data, we vary the subject’s outfit across scenarios. However, due to the higher cost of real-world data collection, the diversity of appearances is inherently lower than in the synthetic setting.
During fine-tuning, we resize the longer side to $1280$ while maintaining the aspect ratio, then apply a center crop of $704 \times 1280$ to match wan2.2~\cite{wan2025wan}.

\section{Limitation and Future Work}
Although \method shows strong performance on egocentric–exocentric video translation, as illustrated in \cref{fig:limitation}, it still exhibits some limitations:
(i) Since \method is fine-tuned over pretrained backbones, its performance ceiling is inherently constrained, especially in scenes involving significant physical changes.
(ii) Our real-world dataset has limited diversity in human subjects. Most of the variation comes from changes in clothing, rather than from distinct individuals. Therefore, the ability of the model to generalize across different characters in egocentric-to-exocentric video translation remains restricted.
Overall, this work provides a new benchmark for egocentric–exocentric video translation, and we leave the development of more generalizable personalized exocentric synthesis to future works.

\section{Additional Qualitative Comparisons}
In addition to qualitative comparisons presented in the main paper, we provide further comparison results here, as illustrated in \cref{fig:additional_qualitative} and \cref{fig:additional_qualitative2}.
Our \method outperforms all baselines on both egocentric-to-exocentric and exocentric-to-egocentric video translation tasks, further highlighting its superior performance.

\begin{figure*}[!t]
    \centering
    \includegraphics[width=0.98\textwidth]{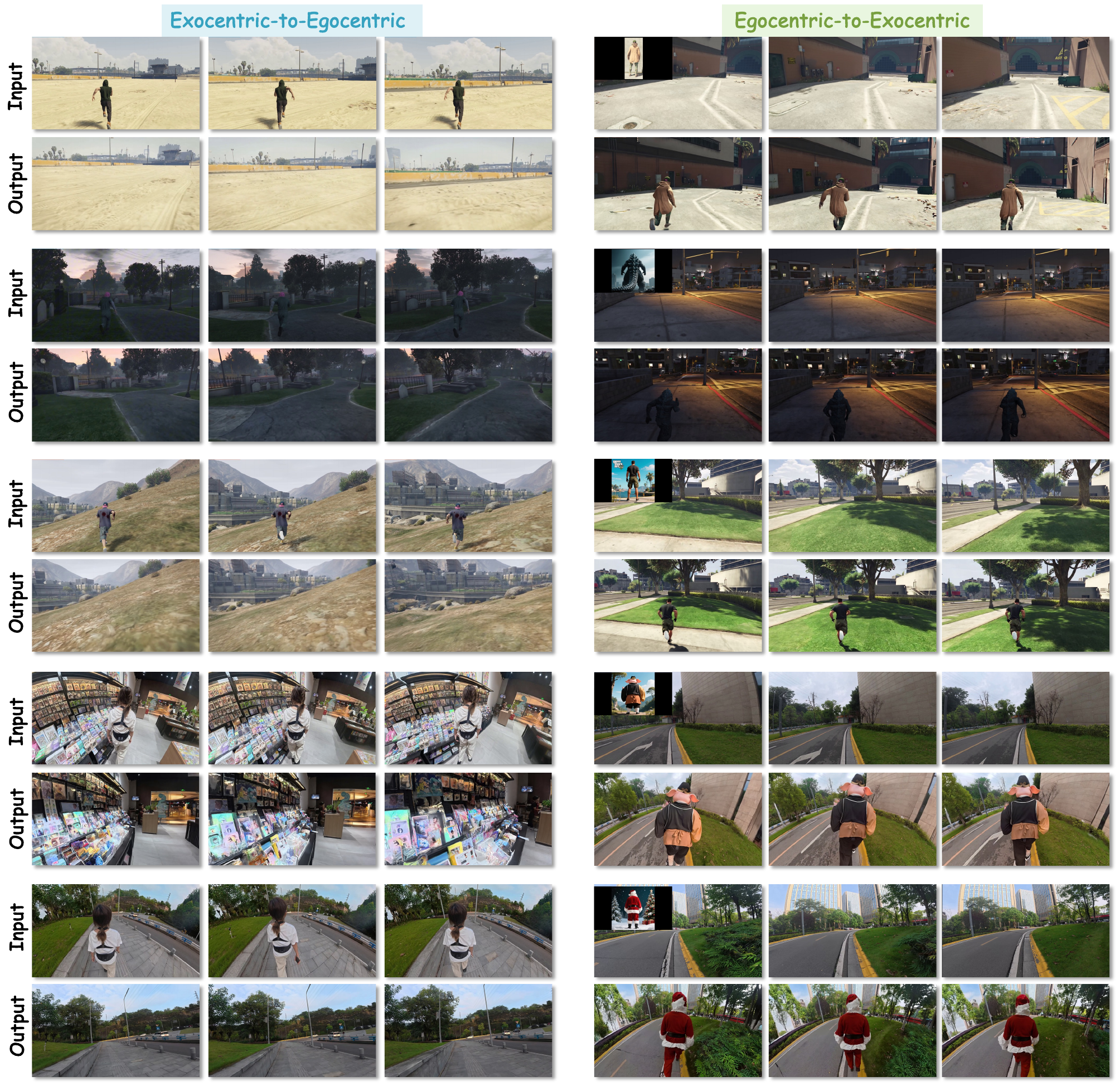}
    \caption{
    Additional gallery of our \method. It supports egocentric-to-exocentric translation for out-of-domain characters. Best viewed with zooming in.
    }
    \label{fig:additional_gallery}
\end{figure*}

\section{Additional Visual Gallery}
To demonstrate the generalization ability of our \method, we provide additional visual gallery in \cref{fig:additional_gallery}.
As shown, our \method supports egocentric-to-exocentric translation for out-of-domain characters, enabling character-centric world exploration.

%
%
\bibliographystyle{splncs04}
\bibliography{main}

\end{document}